\documentclass[conference]{IEEEtran}
\IEEEoverridecommandlockouts
\usepackage{url}
\usepackage{amsmath,amssymb,amsfonts}
\usepackage{algorithmic}
\usepackage{graphicx}
\usepackage{textcomp}
\usepackage{siunitx}
\usepackage{multirow}
\usepackage{subcaption}
\usepackage{xcolor}
\usepackage[numbers,sort&compress]{natbib} 
\def\BibTeX{{\rm B\kern-.05em{\sc i\kern-.025em b}\kern-.08em
    T\kern-.1667em\lower.7ex\hbox{E}\kern-.125emX}}
\begin{document}

\title{NLICE: Synthetic Medical Record Generation for Effective Primary Healthcare Differential Diagnosis
}

\author{
    \IEEEauthorblockN{Zaid Al-Ars$^1$ \hspace{0.4cm} Obinna Agba$^1$ \hspace{0.4cm} Zhuoran Guo$^1$ \hspace{0.4cm} Christiaan Boerkamp$^1$ \hspace{0.4cm} Ziyaad Jaber$^2$ \hspace{0.4cm} Tareq Jaber$^2$}\vspace{0.1cm}
    \IEEEauthorblockA{
    \begin{minipage}{0.49\textwidth}
    \centering
    $^1$Accelerated Big Data Systems\\ Delft University of Technology\\ Delft, The Netherlands
    \end{minipage}
    \begin{minipage}{0.49\textwidth}
    \centering
    $^2$Medvice Digital Health\\ 
    Sint Janssingel 92, 5211DA\\
    's-Hertogenbosch, The Netherlands
    \end{minipage}
    }
}

\maketitle

\begin{abstract}
This paper offers a systematic method for creating medical knowledge-grounded patient records for use in activities involving differential diagnosis. Additionally, an assessment of machine learning models that can differentiate between various conditions based on given symptoms is also provided. We use a public disease-symptom data source called SymCat in combination with Synthea to construct the patients records. In order to increase the expressive nature of the synthetic data, we use a medically-standardized symptom modeling method called NLICE to augment the synthetic data with additional contextual information for each condition. In addition, Naive Bayes and Random Forest models are evaluated and compared on the synthetic data. The paper shows how to successfully construct SymCat-based and NLICE-based datasets. We also show results for the effectiveness of using the datasets to train predictive disease models. The SymCat-based dataset is able to train a Naive Bayes and Random Forest model yielding a 58.8\% and 57.1\% Top-1 accuracy score, respectively.  
In contrast, the NLICE-based dataset improves the results, with a Top-1 accuracy of 82.0\% and Top-5 accuracy values of more than 90\% for both models. Our proposed data generation approach solves a major barrier to the application of artificial intelligence methods in the healthcare domain. Our novel NLICE symptom modeling approach addresses the incomplete and insufficient information problem in the current binary symptom representation approach. The NLICE code is open sourced at \url{https://github.com/guozhuoran918/NLICE}.
\end{abstract}


\section{Introduction}
The process of differential diagnosis is defined as:
\textit{The distinguishing of a disease or condition from others presenting with similar signs and symptoms}.  
In the context of primary healthcare, differential diagnosis describes the process by which a doctor (or other medical practitioners) deduces a possible set of conditions which might be responsible for the symptoms presented by a patient. As a result, the doctor determines the next course of action, such as drug prescription, further testing to rule out disease alternatives, etc. However, despite the ubiquity and importance of this process in the encounter between patients and doctors, it is by no means an easy task. A study~\cite{Singh2014} has shown that in the United States, 1 in 20 outpatient visits is misdiagnosed. Even in cases where a misdiagnosis is not harmful to the patient's health, extra time and financial resources are spent while arriving at the correct diagnosis. This places an extra burden on the patient, the doctor and the healthcare system as a whole. Because of advances in machine learning (ML) techniques~\cite{zhu2020nasb} and computational capabilities~\cite{hoozemans2021fpga}, it should be possible to train ML models capable of supporting the task of obtaining a differential diagnosis given a patient's symptoms. Such models can then be integrated in assistive tools which would, at the very least, confirm the doctor's initial differential diagnosis or suggest possible conditions which might have been overlooked~\cite{diag_suggest}.

However, a limitation in the field of differential diagnosis is the absence of sizable and medically accurate public datasets, due to the sensitive nature of medical information~\cite{stefan2020} and the potential risk of data leakage~\cite{Zaid_fl_attack}. Recent research on automating diagnosis tasks has explored creating synthetic patient records or real-world datasets from free, online symptom checkers. 
Two real-world datasets based on patient self-reports and conversations in a Chinese healthcare online medical forum are Muzhi~\cite{wei-etal-2018-task} and Dxy~\cite{Dxy}. However, due to the limited number of reported symptoms and conditions, these real-world datasets cannot be widely utilized in ML models. One possible approach for expanding the scope of symptoms and conditions is to construct synthetic patient records based on the relationships between conditions and symptoms. Published work~\cite{liao2020task} built synthetic patient datasets by using the SymCat symptom-disease database. Although these real-world datasets and synthetic datasets have the possibility of evaluating the performance of ML models in the automatic diagnosis task, the medical correctness of those datasets cannot be guaranteed because probabilities and statistics used in symptom-condition databases lack professional and reliable medical knowledge. Moreover, the representation of symptoms in current medical records is trivial and incomplete, which results in simplistic symptom-condition assignments (e.g. binary or one-hot vector encoding). This illustrates the need for professional and medically correct datasets in the differential diagnosis research area~\cite{vletter2022towards}.

This paper presents a systematic method to simulate medically-correct and highly-expressive patient records, which integrates a medically standardized symptom modeling approach called NLICE (pronounced as \textit{/en-lais/})~\cite{nlice_git}. NLICE symptom modeling is a novel way to enhance the symptom representation, which is therefore useful for identifying diseases that have similar symptoms. Furthermore, in collaboration with medical experts, medically correct symptom-condition statistics are provided to increase the accuracy of simulated patient records. We generate two types of datasets: one built with only SymCat symptom-condition database, and another with augmented NLICE information. The SymCat dataset includes a total of 5 million records, covering 801 distinct conditions each with 376 potential symptoms. The NLICE dataset uses statistics provided by medical experts, and includes 55 conditions each with 137 potential symptoms. In this way, we can bridge the gap between limited available patient records and data-driven healthcare methodologies.

\section{Modeling of differential diagnosis} \label{sec:diagnosisML}
If we denote all patient information available as $p=P$ and each condition $C_i$ in the set of all possible conditions (hypothesis) $C$ where $C_i \in C  \forall i$ then more formally we can state that the doctor ranks conditions using the probability:

\begin{equation} \label{equation: ranking-prob}
Pr(c=C_i | p=P)
\end{equation}

In colloquial terms, Equation \ref{equation: ranking-prob} gives the probability that the patient's condition is $C_i$ given that $P$ captures all available patient information (e.g. symptoms, demography (sex, age, race), medical history, environmental conditions, etc).

Using Bayes law which can be stated as follows:
\begin{equation}
Pr(Y| X) =  \frac{Pr(X|Y) \times Pr(Y)}{Pr(X)}
\label{equation: bayes-law}
\end{equation}

where $Pr(X)$ and $Pr(Y)$ represent the prior probability distribution of $X$ and $Y$ respectively,  $Pr(Y|X)$ represents the posterior probability and $Pr(X|Y)$ represents the likelihood of $X$ given $Y$, we can then write Equation \ref{equation: ranking-prob} as:

\begin{equation} \label{equation: prob-formulation}
Pr(c=C_i | p=P) = \frac{Pr(p=P | c=C_i) \times Pr(c=C_i)}{Pr(p=P)}
\end{equation}

This Bayesian formulation of the differential diagnosis task can be used as a baseline for comparing the effectiveness of other ML based differential diagnosis methods. 

\section{Methods}
Medical data typically contains sensitive information about the patients, this coupled with an increase in privacy regulations surrounding access and utilization of data makes it especially difficult to access real patient electronic health records.

A number of patient record generators~\cite{emrbots, synthea} have been developed in a bid to address these difficulties. 
In this paper, we use SymCat~\cite{misc-SymCat}, a public symptom-condition data source, in combination with Synthea to generate the data used subsequently for the analysis. The following subsections provide more details regarding the selected data source and generator.

\subsection{Synthea}
Synthea~\cite{synthea}, a synthetic patient population simulator, allows for the generation of \textit{realistic} patient medical records. To avoid privacy concerns, the generator was developed relying on publicly available medical information and health statistics. The project is also fully open-source with a permissive license which allows prospective users to modify the codebase to suit target applications. In its earliest version, Synthea modeled conditions ranked as the top 10 causes of visits to a primary healthcare provider and the top 10 conditions according to the "\textit{years of life lost}" metric. Since this initial version, support has been added for more conditions with many of these contributed by its active community. 

While Synthea does generate patient records, more focus is placed on activities carried out during encounters with healthcare providers (e.g. laboratory procedures, payments, prescribed medication, etc). Little attention is placed on the symptomatic expression of a particular condition. There was no statistical relationship between the condition being modeled and the symptoms presented. In such cases, once a patient contracts the disease, all symptoms associated with that disease would be expressed in the generated patient record. However, 
Synthea's expressive generic module framework provides means to encode a disease-symptom probabilistic data source into Synthea compatible modules.

\subsection{SymCat}

SymCat is \textit{a disease calculator that uses hundreds of thousands of patient records to estimate the probability of disease}~\cite{misc-SymCat}. It provides an interface where users can supply information about the symptoms being experienced and receive a differential diagnosis. In addition, SymCat provides a conditions and symptoms directory. This knowledge base which is publicly available on SymCat's website\footnote{A scrapped CSV version of this data was provided by Alexis Smirnov of  \url{https://www.dialogue.co/en} and is publicly available at \url{https://github.com/teliov/SymCat-to-synthea}} provides probabilistic relationships between 474 symptoms and 801 conditions. 

SymCat data for each condition also contains the gender-based odds of contracting the disease. Also included are race-based odds for disease contraction. SymCat contains 4 race divisions: \textit{White}, \textit{Black}, \textit{Hispanic} and \textit{Others}. Finally, age based odds for contracting each condition are provided. SymCat has 8 age groups: \textit{$<$ 1 year}, \textit{1-4 years}, \textit{5-14 years}, \textit{15-29 years}, \textit{30-44 years}, \textit{45-59 years}, \textit{60-74 years} and \textit{$>$ 75 years}. It should be stated that out of the 474 symptoms, only 376 were associated with a condition. Symptoms with no associated condition were dropped from the SymCat data source.

A more formal description of SymCat's data is given below:
\begin{itemize}
	\item A list of conditions $C$ is provided.
	\item For each condition $C_i$, the age based odds $Pr(c=C_i | a=A_j)$ for each age group $A_j$ are provided.
	\item Also, for each condition, the gender based odds $Pr(c=C_i | g=G_k)$ are provided given that $G_k \in \{male, female\}$.
	\item Race based odds $Pr(c=C_i | r=R_l)$ for each race group $R_l$ for each condition are also provided.
	\item For each condition $C_i$, a set of symptoms $S^i$ which might be presented for that condition along with the probability that the symptom is presented $Pr(S_{m}^{i} | c=C_i)$ where $S_{m}^{i} \in S^{i}$ is also provided.
	\item Additionally, for each symptom $S_i$, age based odds $Pr(s=S_i | a=A_j)$, race based odds $Pr(s=S_i| r=R_l)$ and gender based odds $Pr(s=S_i | g=G_k)$ are also provided.
\end{itemize}
\subsubsection*{Combining SymCat and Synthea}

We developed a Python application\footnote{\url{https://github.com/teliov/SymCat-to-synthea}} in collaboration with Arsène Fansi Tchango\footnote{\url{https://github.com/afansi}} of the MILA\footnote{\url{https://mila.quebec/en/mila/}} research institute in Quebec. The application parses the CSV SymCat data and generates Synthea compatible modules. When generating the Synthea modules, the probability $Pr(c=C_i | a=A_j, g=G_k, r=R_l)$ is first determined using data provided in SymCat. 
This probability can be expressed using Bayes law (Equation \ref{equation: basic-generation-1}) as follows:
\begin{equation} \label{equation: basic-generation-1}
 \begin{split}
     Pr(c=C_i|a=A_j,g=G_k,r=R_l) = \\ \frac{Pr(a=A_i,g=G_k,r=R_l|c=C_i) \times Pr(c=C_i)}{Pr(a=A_i, g=G_k, r=R_l )}
 \end{split}
\end{equation}

In order to simplify the generation process, a conditional independence is assumed between the patient's age, gender and race given the patient's condition. Also assuming that all conditions have the same prior and with a repeated application of Bayes law we obtain:
\begin{equation}
\begin{split}
    Pr(c=C_i | a=A_j, g=G_k, r=R_l) = Pr(c=C_i | a=A_j) \\ \times Pr(c=C_i | g=G_k) \times Pr (c=C_i | r=R_l)
\end{split}
\end{equation}

This gives the probability with which the Synthea generator will allow a patient with age $A_j$, gender $G_k$ and race $R_l$ to contract the disease or condition $C_i$. Once a patient record has been generated by Synthea, the presented symptoms are simply picked based on the probability $Pr(S_{m}^{i} | c=C_i)$ provided by the SymCat data source.

It is worth noting that the conditional independence assumption in data generation deviates from what is obtainable in practice. It is not unusual to have a more complex relationship between the patient demography and the contracted condition as well as the expressed symptoms~\cite{race_diabetes}. Nonetheless, with the data available, this was a reasonable approximation to make.

\subsection{NLICE database}
The SymCat-based database mentioned above offers a machine-usable version of patient medical records: A binary value of 1 denotes the presence of a symptom, while a value of 0 denotes the absence of that symptom. However, we could provide more information about symptom characteristics without extra laboratory or diagnostic procedures. By doing so, we could better describe medical conditions and increase the effectiveness of differential diagnosis.

\subsubsection{NLICE symptom modeling}
Providing more characteristics in the symptom's expression can help make a better differential diagnosis in practice. These characteristics are summarized with the NLICE acronym, which is short for Nature, Location, Intensity, Chronology, and Excitation.

We also use Synthea's expressive generic module framework to generate representative datasets that include this information as long as statistical relationships between these additional symptom characteristics and respective conditions are known. 

\subsubsection * {Nature}

Nature refers to the various ways a specific symptom can manifest itself~\cite{franco2017review}. For example, we use coughing as an illustration, which is a typical symptom connected to respiratory diseases. Modeling this symptom using a one-or-zero approach ignores the various ways coughing may occur. A dry cough could be experienced by some patients as opposed to others who may cough up mucous. These variations specify the type of cough and offer details that might make it simpler to differentiate between conditions that share similar symptoms.

\subsubsection* {Location}

Location refers to the position on the body where a patient experiences a symptom. The location of a symptom can be a discriminating factor when making distinctions between conditions. For example, we consider a patient who reports experiencing abdominal pain. In terms of medical anatomy, the abdomen can be divided into upper and lower right quadrants as well as upper and lower left quadrants~\cite{hepburn2004examination}. Additionally, the abdomen area can also be divided into nine sections, including the epigastric, umbilical and hypogastric regions, right and left hypochondriac, right and left lumbar, and right and left iliac~\cite{bilal2017clinical}. It should be easier to identify the underlying condition if we have more details about which quadrants or parts of the abdomen experience pain.

\subsubsection*{Intensity}

The intensity of a symptom refers to the severity at which a symptom is experienced. 
The severity of the symptom presentation may clearly indicate the underlying illness. For instance, a common symptom of many ailments is pain. An appendicitis sufferer will commonly experience severe to moderate abdominal discomfort. One thing to note here is that intensity is typically a highly subjective experience. A patient's emotional and mental condition, for example, might have a significant impact on their experience~\cite{venkiteswaran2021angry}, making it challenging to precisely measure intensity. 

\subsubsection*{Chronology}

Chronology encompasses three different concepts: 1.~frequency, which refers to how frequently a symptom occurs, 2.~duration, which refers to how long the symptom lasts, and 3.~onset, which describes the time the patient first noticed the symptom. When determining a differential diagnosis, these characteristics may provide diagnostic value.

\subsubsection*{Excitation}

Activities that patients engage in or situations patients are exposed to that activate or worsen the symptoms are referred to as excitation~\cite{lisman2012excitation}. For instance, a patient may only experience heart pain while swimming. We could also distinguish conditions more precisely if we recorded excitation information.

\subsubsection{NLICE data collection}
An NLICE modeling strategy is not directly applicable for the data presented in SymCat. This is despite the fact that there are conditions that presented, for example, with \textit{burning-abdominal} pain and others that presented with \textit{sharp-abdominal} pain, thereby allowing SymCat to capture some elements of the NLICE technique (both these distinctions are made on the nature of the abdominal pain). However, not all symptoms that could support the NLICE strategy are covered by this approach. Therefore, we acquired data for a list of conditions from the medical literature as a proof of concept. This data was collected by our industry partners at Medvice, who are medical specialists. The conditions in our database were divided into ten groups for the purposes of data collection. 

\subsubsection{Combining NLICE and Synthea}

We created Synthea-compatible modules and an adjusted Synthea generator\footnote{\url{https://github.com/guozhuoran918/NLICE}} to generate the dataset, similar to the application we created for SymCat-based databases. The probability of race, gender, and age is provided by the NLICE source data. The first stage of the application is to simulate a patient with the following characteristics:  
According to Bayes law (Equation~\ref{equation: bayes-law}), the probability $Pr(c=C_i | a=A^i_j, g=G_k, r=R_l)$ in the NLICE database is stated as follows:

\begin{equation} \label{equation: basic-generation-2}
 \begin{split}
     Pr(c=C_i|a=A^i_j,g=G_k,r=R_l) = \\ \frac{Pr(a=A^i_j,g=G_k,r=R_l|c=C_i) \times Pr(c=C_i)}{Pr(a=A^i_j, g=G_k, r=R_l )}
 \end{split}
 \end{equation}
We assume conditional independence still holds in this formula, which yields:
\begin{equation}
\begin{split}
    Pr(c=C_i | a=A^i_j, g=G_k, r=R_l) = Pr(c=C_i | a=A^i_j) \\ \times Pr(c=C_i | g=G_k) \times Pr (c=C_i | r=R_l)
\end{split}
\end{equation}

where $A^i_j$ stands for the probability of age group $j$ in condition $i$.

Subsequently, the Synthea generator picks the presented symptoms according to the probability $Pr(S_{m}^{i} | c=C_i)$ provided by the NLICE data source. At the same time, each presented symptom will be associated with eight NLICE features: $[symptom:nature:location:intensity:frequency:duration:onset:excitation]$ (where frequency, duration and onset encode chronology). However, each condition does not need to contain all NLICE features. The adjusted Synthea generator displays only those conditions for which "NLICE" is known.
For example, if there is no data about the \emph{nature} of (e.g. in the case of the fatigue present in covid-19) then the NLICE data point \emph{nature} is not displayed.

\subsection{Selected ML models}
For this problem, we select two popular ML models often employed in the medical domain: naive Bayes and random forest.

\subsubsection*{Naive Bayes}

The Naive Bayes model is the model we use as a baseline, given the probabilistic problem formulation. 
The conditional independence assumption used in the data generation process is the same assumption made by the Naive Bayes algorithm. Despite this strong assumption, Naive Bayes still achieves reasonable accuracy~\cite{nb-optimal, nb-zhang}. Also, this model allows for the most suitable probability distribution function to be used for each feature. This flexibility allows us to assume that the patient's age is distributed according to a Gaussian distribution, the patient's race assumes a categorical probability distribution and the gender along with all the symptoms (which take on values of 0 or 1) assumes a Bernoulli distribution.

\subsubsection*{Random Forest}

The suitability of a Random Forest model is also evaluated. Random Forest is an ensemble method that uses bagging to combine predictions from a collection of decision trees each trained on a random subset of features~\cite{decision-tree}. It has been shown to be very robust to noise and also avoids the over-fitting commonly associated with single decision trees~\cite{random-forest}. With very few parameters for optimization and the nature of our problem (decision-making), this model is also a good alternative.

\subsection{Evaluation metrics}

Three metrics were selected on which the models would be evaluated: Top-1 accuracy, precision and Top-5 accuracy.

\subsubsection*{Top-1 Accuracy}

At its core, the differential diagnosis task has been formulated as a multi-class classification problem. This makes the Top-1 accuracy a reasonable metric to evaluate the models on.

\subsubsection*{Precision}

Model precision i.e. $\frac{tp}{tp+fp}$ is also considered as an evaluation metric. While precision is more widely associated with binary classification problems, a multi-class extension~\cite{scikit-learn} was employed to allow for reporting a single precision value for all classes.

\subsubsection*{Top-5 Accuracy}

For this task, the differential diagnosis was taken as the first 5 predictions made by the models. As a result, the Top-5 accuracy metric was also considered. A prediction is considered to be Top-5 accurate if the correct condition is one of the most probable 5 model predictions.

\section{Mimicking real-world scenarios}

The use of synthetic data naturally raises the question: \textit{how will the models behave in a real-world setting?}. In an attempt to answer this question, three scenarios were considered for evaluation.

\subsection{Varying minimum number of symptoms per condition}

When generating the baseline data, patients were allowed to contract a condition and express only one symptom. While this is not an impossible scenario, it is more likely that, for any condition which a patient suffers from, more than one symptom would be presented. Hence, evaluation datasets were generated for which the minimum number of symptoms expressed per condition was varied from 2 to 5.

\subsection{Perturbing the condition-symptom probabilities}

Real-world datasets are expected to deviate from the synthetic dataset in terms of the underlying distribution which models the relationship between the patient, conditions and symptoms. Restricting this deviation to the condition-symptom expression probabilities, we can observe how the models would perform when evaluated on data generated with these expression probabilities perturbed.

In implementing this scenario, given a perturbation percentage $\delta$ and the expression probability of a symptom $S_j$ for condition $C_i$, then the perturbed expression probability is obtained as $1 \pm \delta$. The choice to apply $\delta$ in an additive or subtractive manner is randomized. $\delta$ is selected from the set $\{0.1, 0.2, 0.3, 0.5, 0.7\}$.

\subsection{Injecting additional symptoms}

Another realistic scenario is one where the set of symptoms $S^{i}$ associated with a condition $C_i$ is only a subset of the actual set of symptoms associated with that condition. This implies that in such a setting, there might exist other symptoms $S^{m}$ related to condition $C_i$ for which $S^{i} \cap S^{m} = \emptyset$.

To simulate this scenario, a maximum of 5 symptoms are injected into the symptom expression list for condition $C_i$. The new symptoms are selected based on a \textit{similarity} measure $K$ with existing symptoms. To define this similarity measure we adopt a graphical view of symptoms and conditions with symptoms representing nodes and conditions representing edges. Hence, given a condition $C_i$ with its set of symptoms $S^i$ and given that $S^m$ represents the set of symptoms such that $S^i \cap S^m= \emptyset$ then the likelihood of a symptom $S_k^m$ where $S_k^m \in S^m$ being presented by the condition $C_i$ is given as:
\begin{equation}
 	K = \sum_{j=1}^{i} E_{ki}
\end{equation}
where $E_{ki}$ is the edge count between symptom $S_j^i$ and $S_k^m$ given that $S_j^i \in S^i$ and $S_k^m \in S^m$. In essence, $K$ measures how often the symptom being considered is presented in other conditions $C_j \mid j \neq i$ alongside symptoms of the condition $C_i$. A higher $K$ value is taken to indicate a higher similarity with existing symptoms of $C_i$.

Once the 5 most \textit{similar} symptoms have been identified, they are assigned expression probabilities in three methods: the minimum, maximum and mean expression probability for existing symptoms.

\section{Results for baseline synthetic data}


\subsection{Evaluation metric scores} Both the Naive Bayes and Random Forest models were trained and evaluated on the baseline data. Three metrics were selected on which the models would be evaluated: Top-1 accuracy, precision and Top-5 accuracy. Tab.~\ref{table:baseline-perf} shows the results of both models using the selected evaluation metrics in SymCat and NLICE datasets.

\begin{table*}[htbp]
\centering
\caption{Results of models on baseline data}
\begin{tabular}{|l|ll|ll|ll|}
\hline
\multirow{2}{*}{Model} & \multicolumn{2}{l|}{Top-1}          & \multicolumn{2}{l|}{Precision}      & \multicolumn{2}{l|}{Top-5}          \\ \cline{2-7} 
                       & \multicolumn{1}{l|}{SymCat} & NLICE & \multicolumn{1}{l|}{SymCat} & NLICE & \multicolumn{1}{l|}{SymCat} & NLICE \\ \hline
Naive Bayes   & \multicolumn{1}{l|}{0.588} & 0.820 & \multicolumn{1}{l|}{0.633} & 0.845 & \multicolumn{1}{l|}{0.853} & 0.975 \\ \hline
Random Forest & \multicolumn{1}{l|}{0.571} & 0.820 & \multicolumn{1}{l|}{0.612} & 0.902 & \multicolumn{1}{l|}{0.845} & 0.990 \\ \hline
\end{tabular}
\label{table:baseline-perf}
\end{table*}

Tab.~\ref{table:baseline-perf} shows that the Naive Bayes model slightly outperforms Random Forest in the SymCat-based dataset, while Random Forest performs better in the NLICE-based dataset. We also notice the relatively low SymCat Top-1 and Top-5 accuracy scores compared to NLICE which underscores the diagnostic value of augmenting the dataset with extra symptom characteristics. This increase in accuracy can be attributed to a number of factors. One is the similarity in symptom expression in SymCat for similar conditions, which makes an accurate diagnosis very difficult if not impossible for ML models. 
Another factor is the number of symptoms expressed per condition. As mentioned earlier, the baseline dataset is allowed to have patients with conditions presenting only one symptom. In such a case, it would be impossible to distinguish conditions based on only one such symptom. 

\subsection{Qualitative evaluation} For a qualitative evaluation, we observe the predictions made by the models. Due to the large number of conditions in the dataset being evaluated, we select Asthma condition as a case study. Two evaluations are carried out. Tab.~\ref{table: eval-resp-top5} shows the 5 conditions which the models most often misclassify the selected condition as. This is a sort of qualitative confusion matrix. We see from the results that in all cases for Asthma predictions, the \textit{confusion} in the SymCat-based dataset is mostly due to other respiratory conditions. This reinforces the reasoning that the models are unable to distinguish between similar conditions based on symptoms alone in SymCat. In contrast when using NLICE, Asthma is not misclassified as other similar respiratory conditions. This indicates that modeling symptoms with NLICE characteristics can be helpful to distinguish the conditions that contain similar symptoms.

\begin{table*}[tb]
\centering
\caption{Top 5 commonly included predictions for Asthma }
\label{table: eval-resp-top5}
\begin{tabular}{|l|ll|ll|}
\hline
\multirow{2}{*}{Condition} & \multicolumn{2}{l|}{Naive Bayes}                             & \multicolumn{2}{l|}{Random Forest}                           \\ \cline{2-5} 
                           & \multicolumn{1}{l|}{SymCat}             & NLICE              & \multicolumn{1}{l|}{SymCat}             & NLICE              \\ \hline
Asthma                     & \multicolumn{1}{l|}{Acute bronchospasm} & Breast cancer      & \multicolumn{1}{l|}{COPD}               & Breast cancer      \\
                           & \multicolumn{1}{l|}{COPD}               & Otitis externa     & \multicolumn{1}{l|}{Acute bronchospasm} & Otitis externa     \\
 & \multicolumn{1}{l|}{ARDS}                & Lower urinary tract infection & \multicolumn{1}{l|}{ARDS}                & Lower urinary tract infection \\
                           & \multicolumn{1}{l|}{Croup}              & Otitis media acuta & \multicolumn{1}{l|}{Croup}              & Otitis media acuta \\
 & \multicolumn{1}{l|}{Acute bronchiolitis} & Atrophic gastritis            & \multicolumn{1}{l|}{Acute bronchiolitis} & Atrophic gastritis            \\ \hline
\end{tabular}
\end{table*}

\subsection{Posterior estimate comparisons}\label{paragraph:posterior} Besides the qualitative comparisons made in the previous discussion, we also observe the posterior estimates for each of the models. As Equation~\ref{equation: bayes-law} states, given an instance, trained ML models can estimate the probability for each class that the instance belongs to this class. These probabilities are called the posterior probabilities, then this instance is classified to the class with the highest posterior probability~\cite{confidence}.
Since the differential diagnosis is usually taken as the Top-5 condition based on the posterior estimates from each model, we can also say that the posterior estimates tell how \textit{confident} the models are about the diagnosis.

Four cases were evaluated: Top-1 accurate, Top-5 accurate, Non Top-1 accurate and Non Top-5 accurate. Top-n Accurate is the posterior probability estimate for both models when they are accurate in predicting the correct diagnosis as the first n conditions. In contrast, Non Top-n Accurate stands for posterior probability estimate when they fail to predict the correct diagnosis as the first n conditions. 

\begin{figure}[htb]
	\centering
	\hspace*{-0.2cm}\begin{subfigure}[!h]{0.26\textwidth}
		\centering
    	\includegraphics[width=\textwidth]{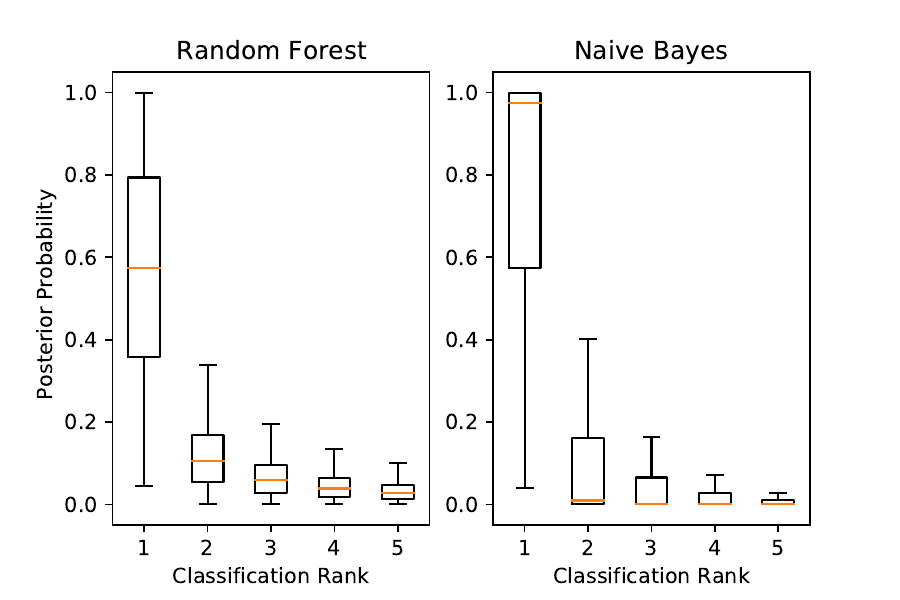}
    	\caption{
    		Top-1}
    	\label{figure: top-1-compare}
	\end{subfigure}
	\hspace*{-0.5cm}\begin{subfigure}[!h]{0.26\textwidth}
		\centering
    	\includegraphics[width=\textwidth]{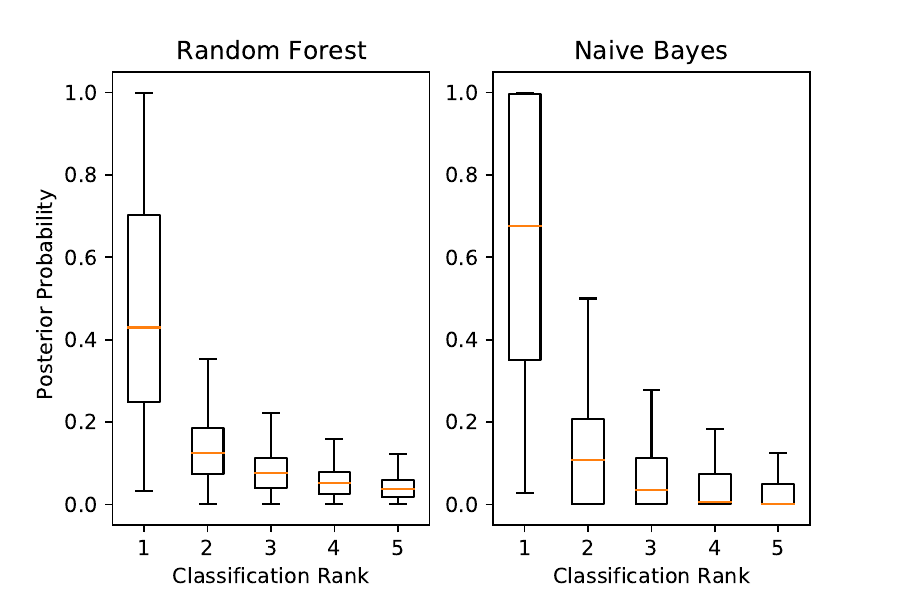}
    	\caption{
    		Top-5
    	}
    	\label{figure: top-5-compare}
	\end{subfigure}
	\vfill
	\hspace*{-0.2cm}\begin{subfigure}[!h]{0.26\textwidth}
		\centering
    	\includegraphics[width=\textwidth]{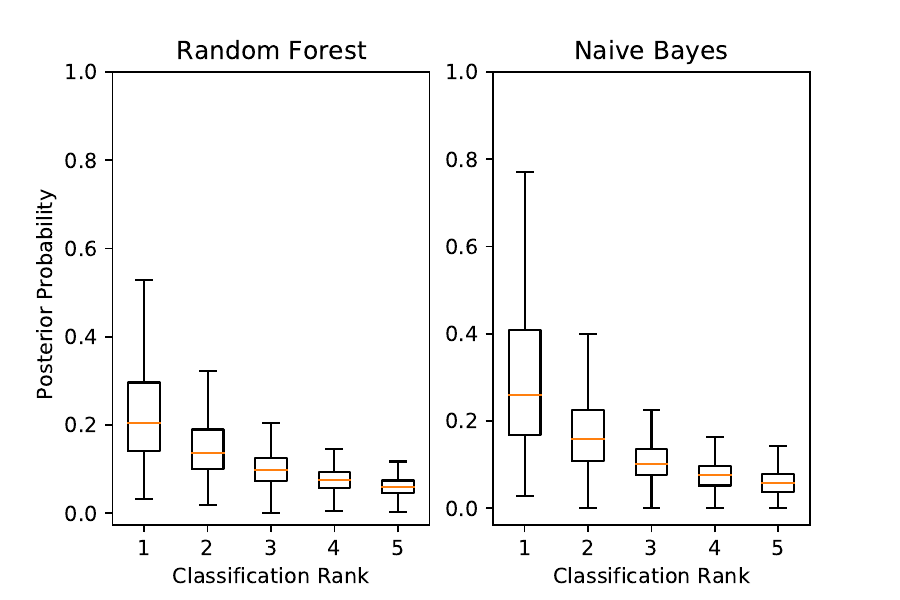}
    	\caption{
    		Non Top-1
    	}
    	\label{figure: top-0-compare}
	\end{subfigure}
	\hspace*{-0.5cm}\begin{subfigure}[!h]{0.26\textwidth}
		\centering
    	\includegraphics[width=\textwidth]{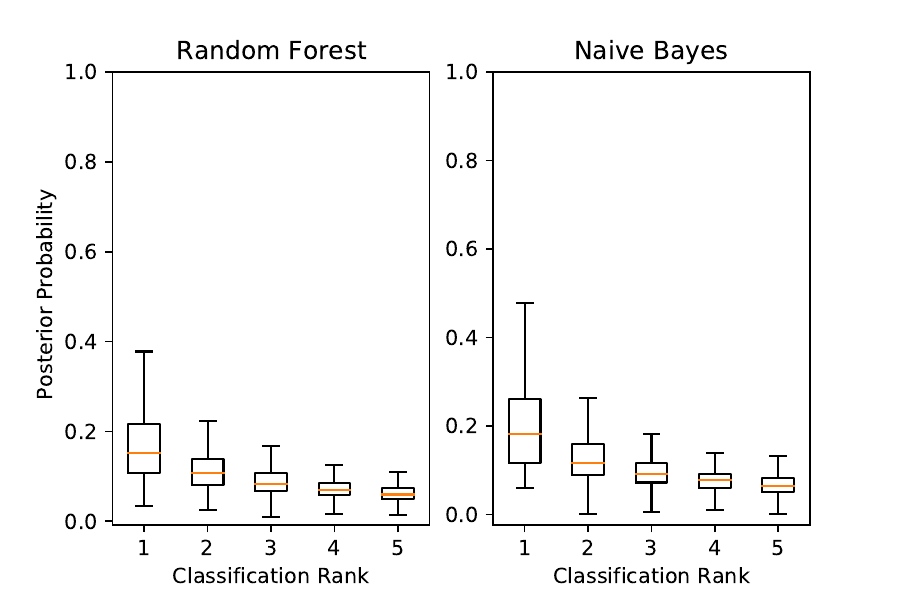}
    	\caption{
    		Non Top-5
    	}
    	\label{figure: top-n5-compare}
	\end{subfigure}
	
	\caption{
	Posterior probability estimate comparisons using Random Forest and Naive Bayes for the SymCat dataset 
	}
	\label{figure: posterior-comp}
\end{figure}

\begin{figure}[htb]
	\centering
	\hspace*{-0.2cm}\begin{subfigure}[h]{0.26\textwidth}
		\centering
    	\includegraphics[width=\textwidth]{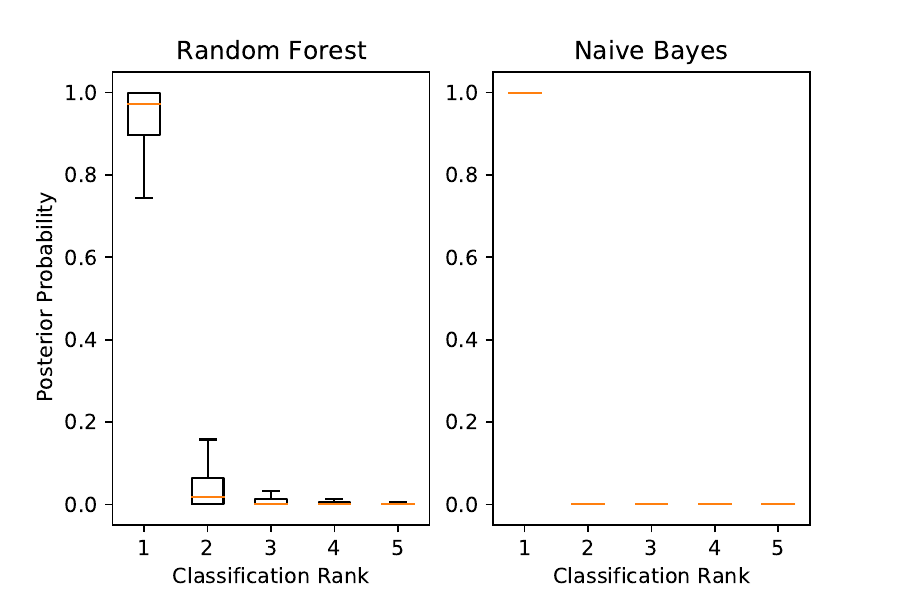}
    	\caption{
    		Top-1}
    	\label{figure: top-1-compare-nlice}
	\end{subfigure}
	\hspace*{-0.5cm}\begin{subfigure}[h]{0.26\textwidth}
		\centering
    	\includegraphics[width=\textwidth]{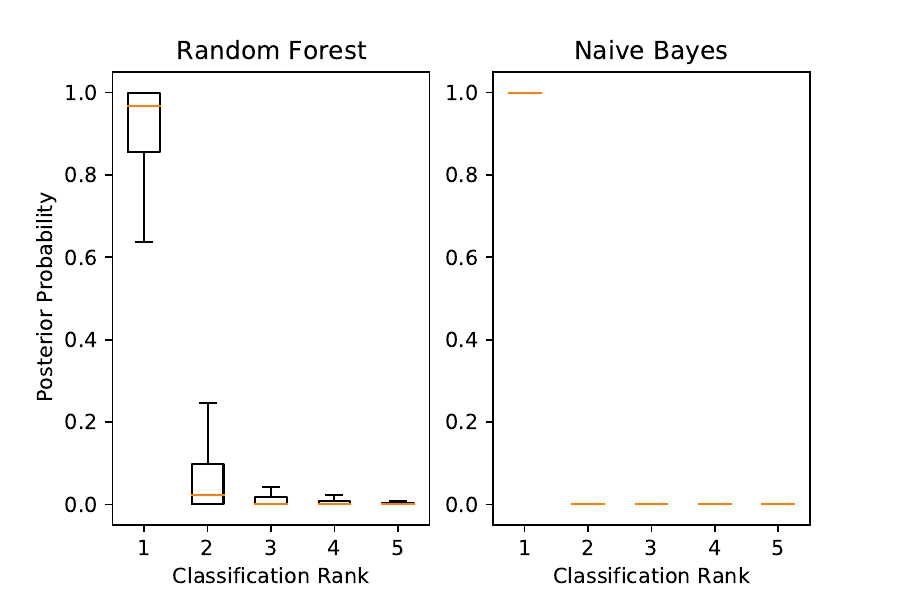}
    	\caption{
    		Top-5
    	}
    	\label{figure: top-5-compare-nlice}
	\end{subfigure}
	\vfill
	\hspace*{-0.2cm}\begin{subfigure}[h]{0.26\textwidth}
		\centering
    	\includegraphics[width=\textwidth]{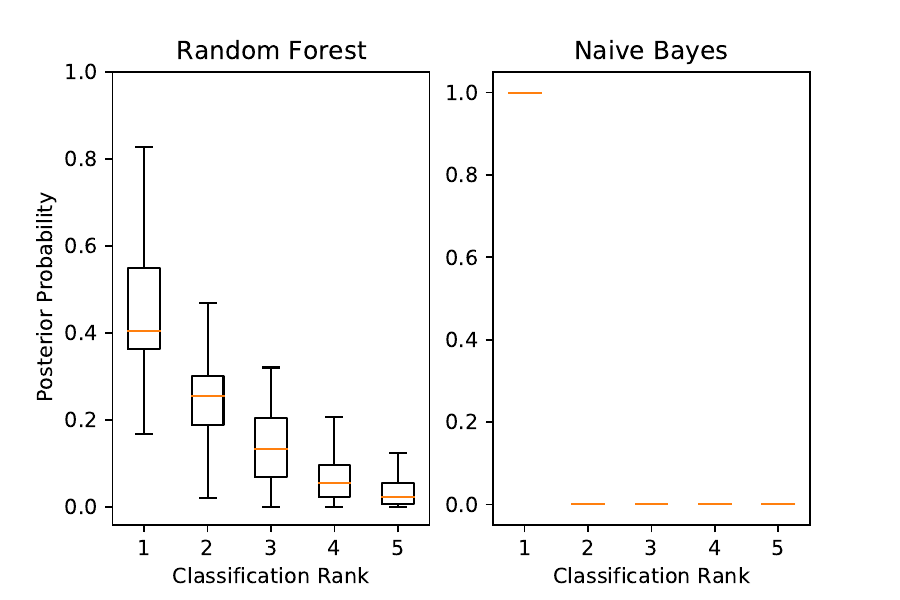}
    	\caption{
    		Non Top-1
    	}
    	\label{figure: top-0-compare-nlice}
	\end{subfigure}
	\hspace*{-0.5cm}\begin{subfigure}[h]{0.26\textwidth}
		\centering
    	\includegraphics[width=\textwidth]{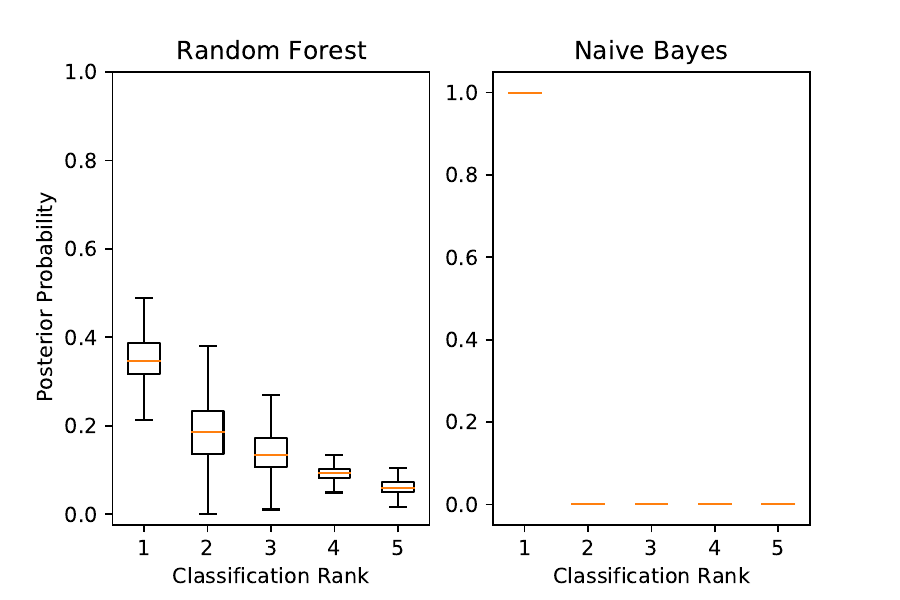}
    	\caption{
    		Non Top-5
    	}
    	\label{figure: top-n5-compare-nlice}
	\end{subfigure}
	
	\caption{
	Posterior probability estimate comparisons using Random Forest and Naive Bayes for the NLICE dataset
	}
	\label{figure: posterior-comp-nlice}
\end{figure}

Fig.~\ref{figure: posterior-comp} shows the experimental results in the SymCat dataset. Naive Bayes has very high Top-1 probability estimates (median of 0.97), when compared to the more moderate Random Forest (median of 0.57). While the median confidence level of Naive Bayes is reduced by 30.52\%, and a 25.28\% reduction for the Random Forest in the Top-5 case. Fig.~\ref{figure: posterior-comp-nlice} shows that same findings hold for the NLICE-based dataset. These results show that Naive Bayes assigns a higher probability estimate to its predicted class compared to Random Forest, which is a well-known behavior~\cite{nb-poor-posterior}. For both models in the SymCat dataset, however, when their predictions are wrong, the probability estimates are much lower as can be seen in Figures \ref{figure: top-0-compare} and \ref{figure: top-n5-compare}. Similarly, the probability estimates for Random Forest in the NLICE dataset are much lower compared to Top-n accurate as shown in Figures \ref{figure: top-0-compare-nlice} and \ref{figure: top-n5-compare-nlice}. This behavior is reasonable because it would be desirable that the models are not very confident when they wrongly predict conditions. However, Naive Bayes yields an overconfident result (either 1 or 0) for almost every sample in the NLICE dataset. One possible explanation is that Naive Bayes fails to learn the complicated relationships between NLICE attributes because it inherently assumes conditional independence between NLICE attributes~\cite{moore2008trouble}, 
which is not the case in our context.

\subsection{Prediction confidence threshold} In a clinical setting, it is important to indicate when the predictions are not confident enough to make clinical decisions. As the results in the previous section indicate, models tend not to be confident when their predictions are wrong, which allows for the use of a confidence threshold in practice~\cite{corbiere2019addressing}. In such a setting, predicted conditions would be presented to the doctor only if the most accurate prediction exceeds the set confidence threshold. As an example, the confidence threshold is selected to be 25th, 50th and 75th percentile as well as the mean of the highest posterior estimate for the model's predictions. 

\begin{figure}[htb]
	\centering
	\hspace*{-0.2cm}\begin{subfigure}[b]{0.26\textwidth}
		\centering
    	\includegraphics[width=\textwidth]{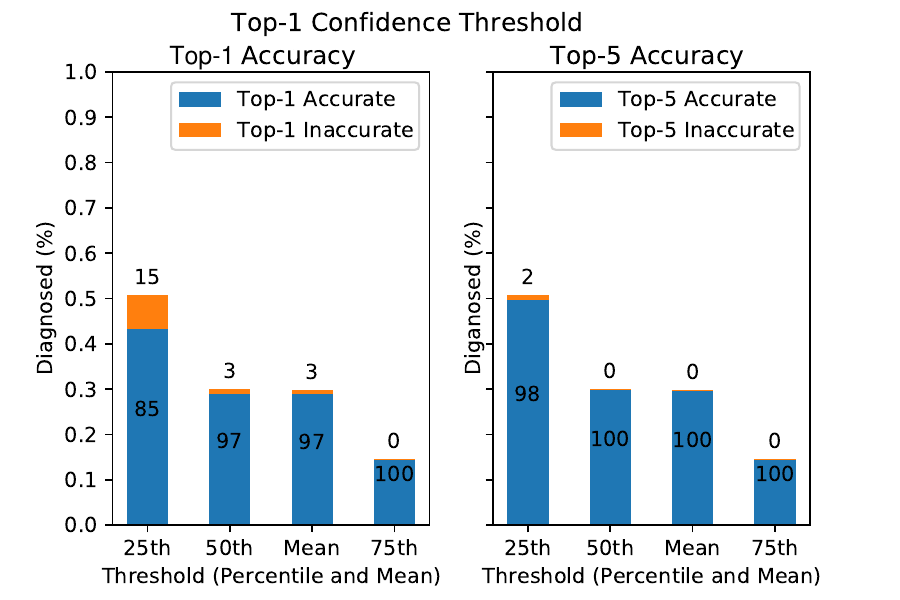}
    	\caption{
    		Top-1
    	}
    	\label{figure: rf-conf-top1}
	\end{subfigure}
	\hspace*{-0.5cm}\begin{subfigure}[b]{0.26\textwidth}
		\centering
    	\includegraphics[width=\textwidth]{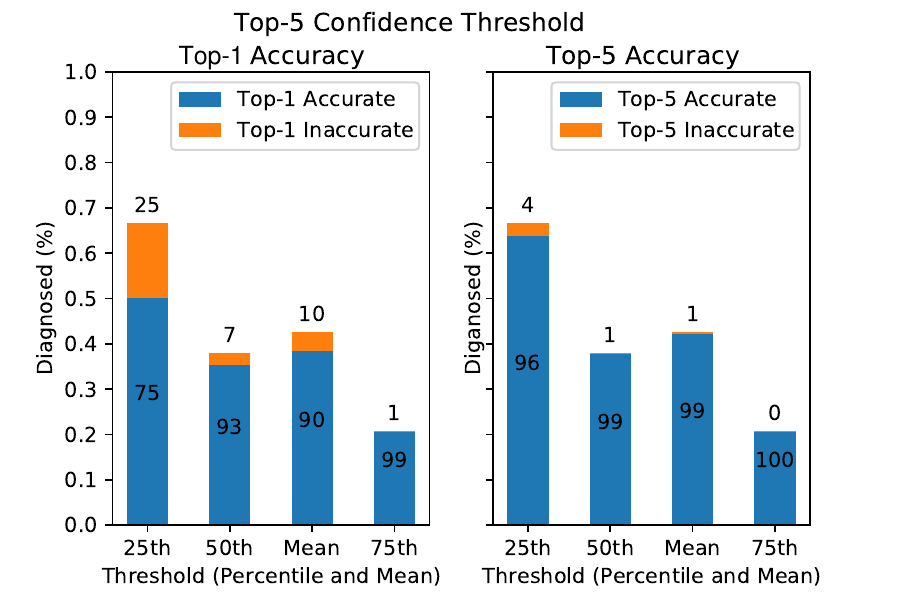}
    	\caption{
    		Top-5
    	}
    	\label{figure: rf-conf-top5}
	\end{subfigure}\\
	\hspace*{-0.2cm}\begin{subfigure}[b]{0.26\textwidth}
		\centering
    	\includegraphics[width=\textwidth]{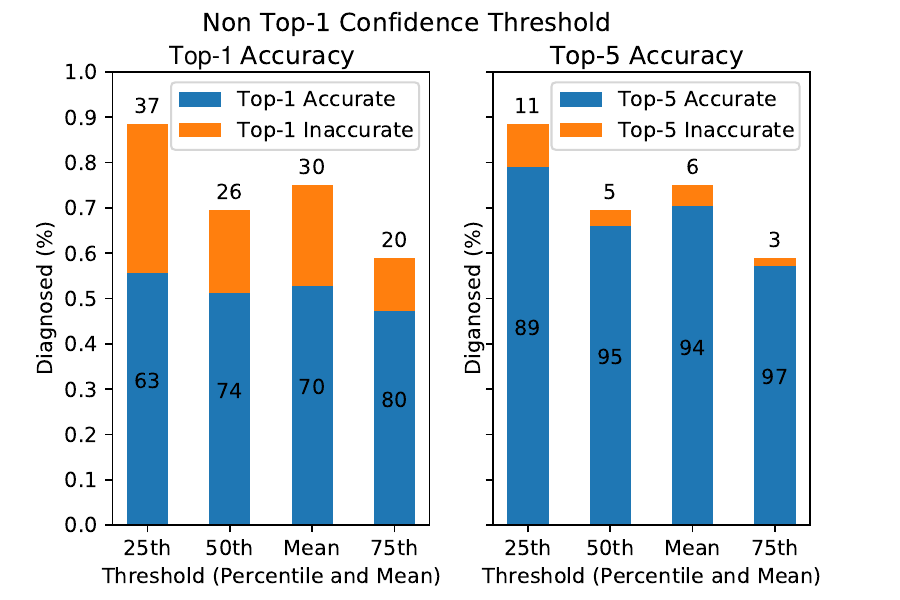}
    	\caption{
    		Non Top-1
    	}
    	\label{figure: rf-conf-top0}
	\end{subfigure}
	\hspace*{-0.5cm}\begin{subfigure}[b]{0.26\textwidth}
		\centering
    	\includegraphics[width=\textwidth]{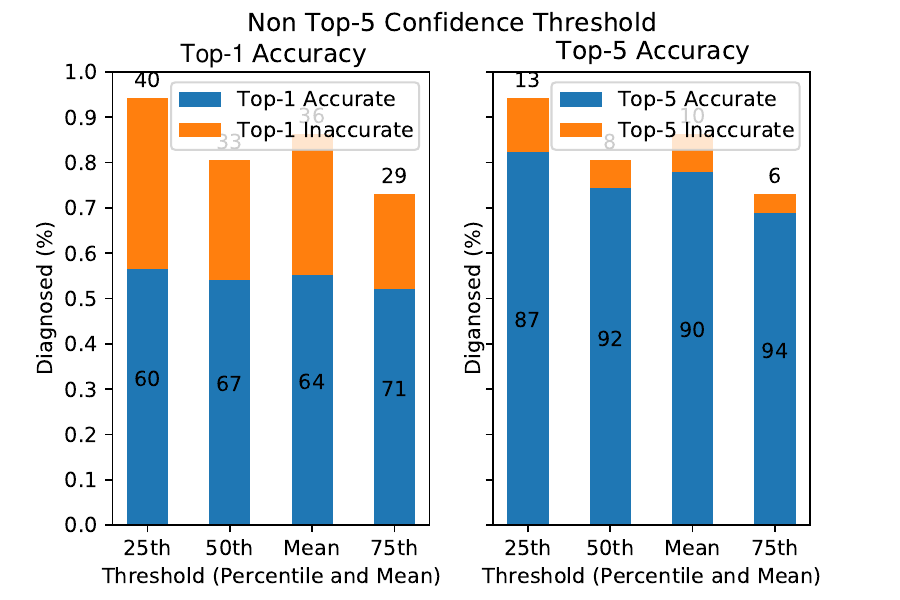}
    	\caption{
    		Non Top-5
    	}
    	\label{figure: rf-conf-topn5}
	\end{subfigure}
	
	\caption[Prediction confidence threshold using Random Forest for the SymCat dataset]{
	Prediction confidence threshold using Random Forest for the SymCat dataset 
	}
	\label{figure: rf-confidence}
\end{figure}

Fig.~\ref{figure: rf-confidence} shows the application of these thresholds to the predictions of Random Forest for the SymCat dataset. We use the same cases mentioned in Section~\ref{paragraph:posterior}: Top-1 accurate, Top-5 accurate, Non Top-1 accurate and Non Top-5 accurate. We calculate the 25th, 50th and 75th percentile as well as the mean of the Top-5 posterior estimate probabilities for each case as threshold values. Subsequently, those threshold values are used to re-evaluate the prediction results in the whole sample tests. In addition, the figure shows the percentage of conditions considered as confident depending on the threshold being used. It also shows what percentage of the confident predictions are accurate. Higher threshold values result in fewer \textit{confident} predictions but also higher accuracy. As can be seen in Fig.~\ref{figure: rf-conf-top1} for Top-1 accurate predictions, using the 25th percentile as a threshold admits approximately half of the predictions as confident (approx.\ 50\% of diagnosed conditions) but has a Top-1 accuracy of 85\% and a Top-5 accuracy of 98\%. Taking the 25th percentile of the Non Top-1 accurate case, the model obtains a 63\% Top-1 accuracy and a 89\% Top-5 accuracy and it admits 88\% of the predictions. In a deployed environment, such a threshold value might be considered reasonable. Predictions below the threshold would not typically be presented to the doctor.

\begin{figure}[htb]
	\centering
	\hspace*{-0.2cm}\begin{subfigure}[b]{0.26\textwidth}
		\centering
    	\includegraphics[width=\textwidth]{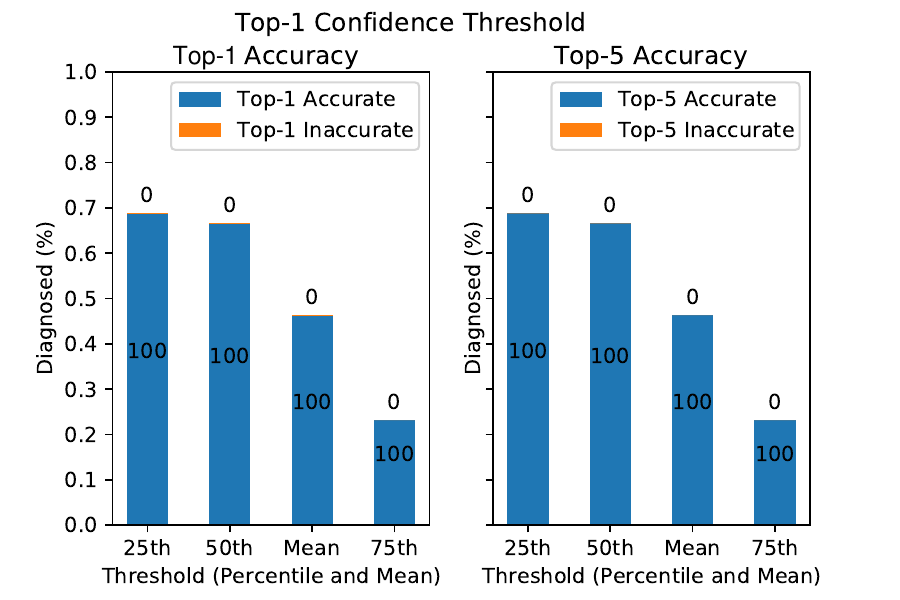}
    	\caption{
    		Top-1
    	}
    	\label{figure: rf-conf-top1-nlice}
	\end{subfigure}
	\hspace*{-0.5cm}\begin{subfigure}[b]{0.26\textwidth}
		\centering
    	\includegraphics[width=\textwidth]{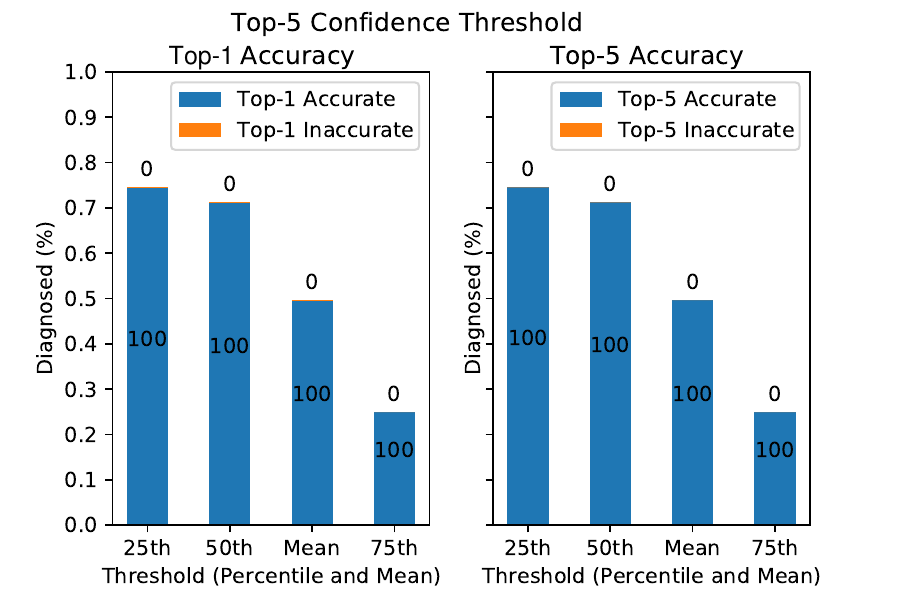}
    	\caption{
    		Top-5
    	}
    	\label{figure: rf-conf-top5-nlice}
	\end{subfigure}\\
	\hspace*{-0.2cm}\begin{subfigure}[b]{0.26\textwidth}
		\centering
    	\includegraphics[width=\textwidth]{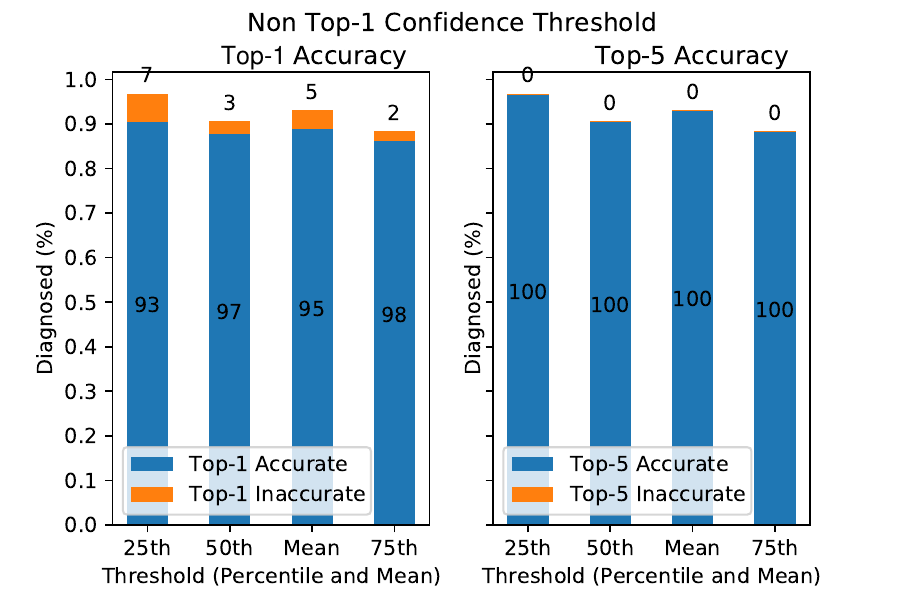}
    	\caption{
    		Non Top-1
    	}
    	\label{figure: rf-conf-top0-nlice}
	\end{subfigure}
	\hspace*{-0.5cm}\begin{subfigure}[b]{0.26\textwidth}
		\centering
    	\includegraphics[width=\textwidth]{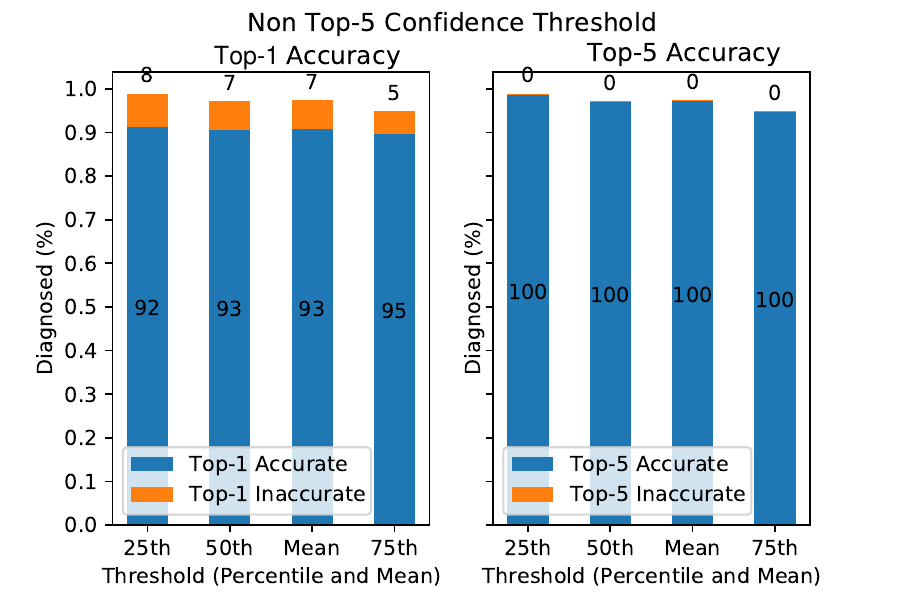}
    	\caption{
    		Non Top-5
    	}
    	\label{figure: rf-conf-topn5-nlice}
	\end{subfigure}
	
	\caption[Prediction confidence threshold using Random Forest for the NLICE dataset]{
	Prediction confidence threshold using Random Forest for the NLICE dataset}
	\label{figure: rf-confidence-nlice}
\end{figure}

Fig.~\ref{figure: rf-confidence-nlice} represents the Random Forest results for the NLICE dataset. The figure shows a significant difference compared to the SymCat dataset: NLICE Top-1 threshold results in Fig.~\ref{figure: rf-conf-top1-nlice} and Top-5 threshold results in Fig.~\ref{figure: rf-conf-top5-nlice} show that increasing the threshold values reduces the number of confident predictions. Non Top-n threshold results in Fig.~\ref{figure: rf-conf-top0-nlice} and Fig.~\ref{figure: rf-conf-topn5-nlice} slightly reduce Top-1 accuracy as the threshold confidence increases. However, most of the predicted conditions by both models can achieve around 100\% accuracy. And the diagnosed percentages in the NLICE dataset are higher than the SymCat dataset, indicating that NLICE models are more confident about their predictions than SymCat models. Applying Non Top-n threshold values in NLICE models admits around 90\% of predictions in the 75th percentile, which means that most Top-1 conditions are assigned a 75th percentile probability in all conditions.

\section{Results for realistic scenarios data}



\subsection{Varying minimum symptoms per condition}
\begin{figure}[ht]
	\centering
	\begin{subfigure}[b]{0.495\textwidth}
		\centering
    	\includegraphics[height=5cm]{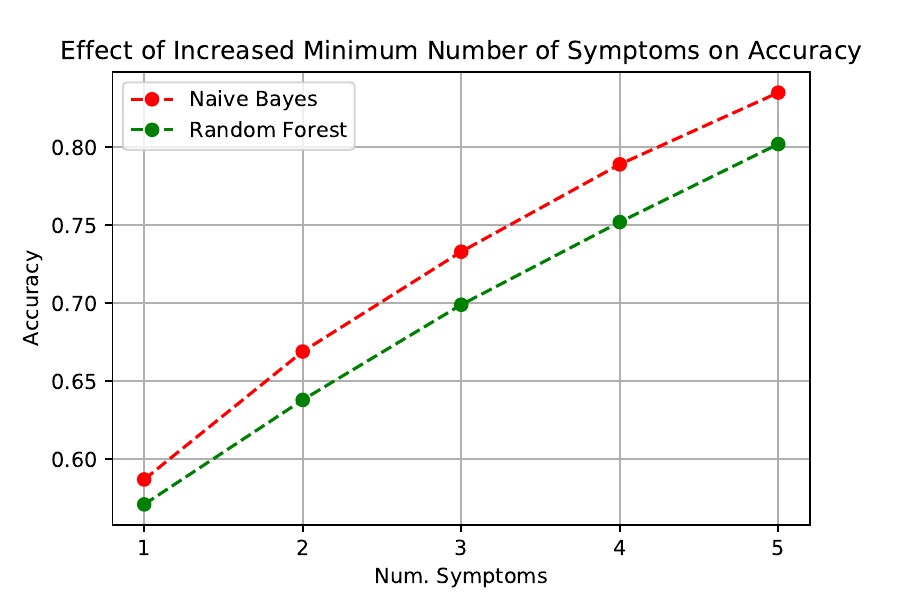}
    	\caption{
    		Top-1 Accuracy
    	}
    	\label{figure: inc-symptoms-acc_symcat}
	\end{subfigure}
	\hfill
	\begin{subfigure}[b]{0.495\textwidth}
		\centering
    	\includegraphics[height=5cm]{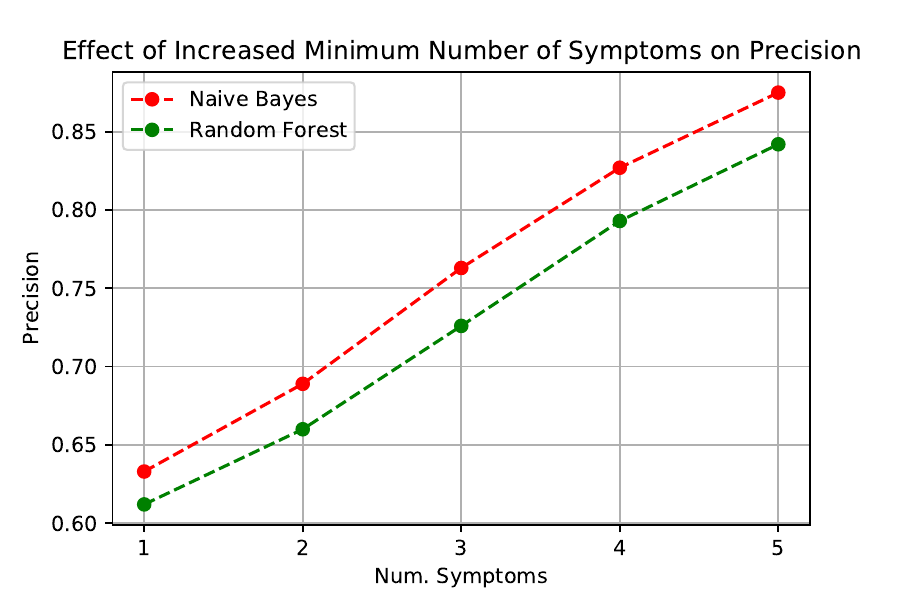}
    	\caption{
    		Precision
    	}
    	\label{figure: inc-symptoms-prec_symcat}
	\end{subfigure}
	\caption[Effect of increased minimum symptom expressed per condition on model performance for SymCat data]{
	Effect of increased minimum symptom expressed per condition on model performance for SymCat data
	}
	\label{figure: inc-symptoms}
\end{figure}

\begin{figure}[ht]
	\centering
	\begin{subfigure}[b]{0.495\textwidth}
		\centering
    	\includegraphics[height=5cm]{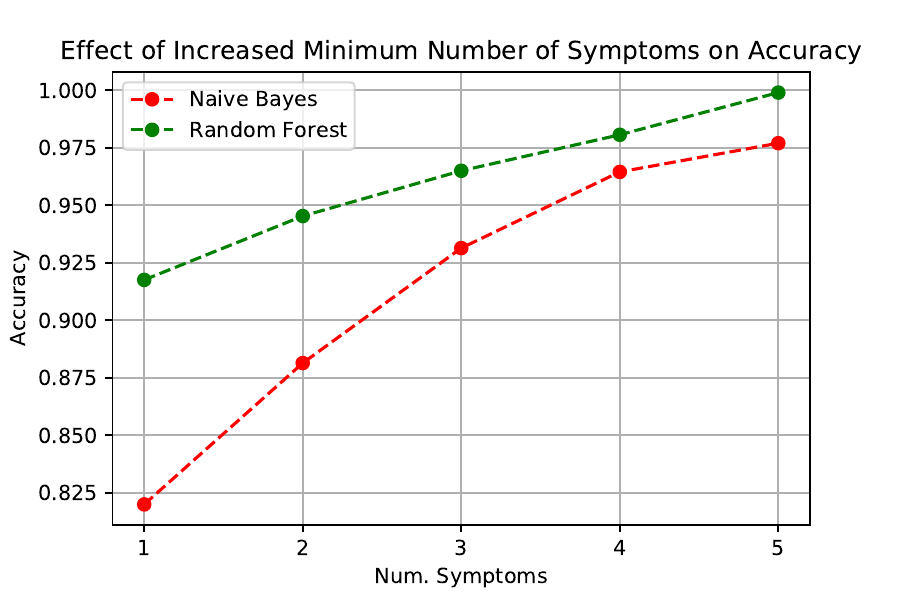}
    	\caption{
    		Top-1 Accuracy
    	}
    	\label{figure: inc-symptoms-acc_nlice}
	\end{subfigure}
	\hfill
	\begin{subfigure}[b]{0.495\textwidth}
		\centering
    	\includegraphics[height=5cm]{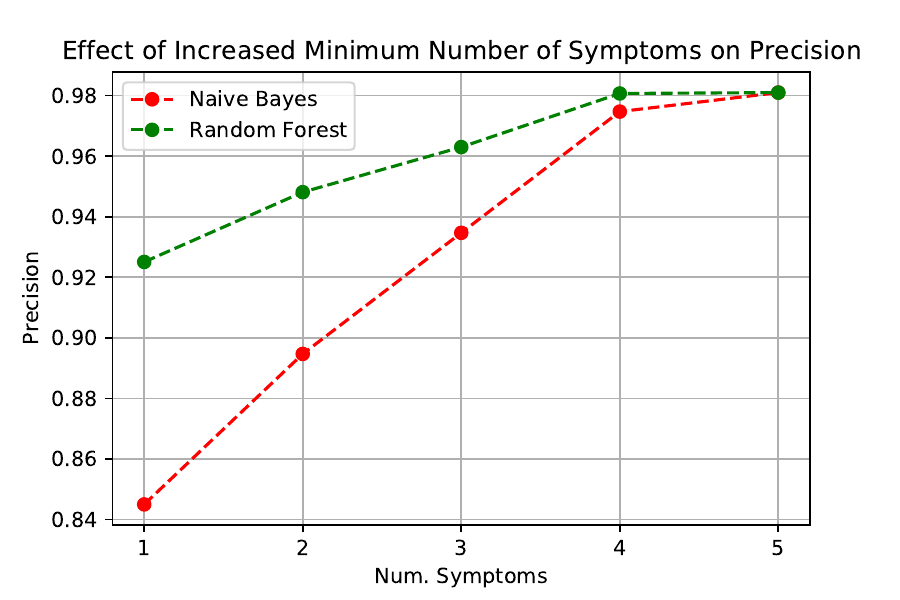}
    	\caption{
    		Precision
    	}
    	\label{figure: inc-symptoms-prec_nlice}
	\end{subfigure}
	
	\caption[Effect of increased minimum symptom expressed per condition on model performance for NLICE data]{
	Effect of increased minimum symptom expressed per condition on model performance for NLICE data
	}
	\label{figure: inc-symptoms-nlice}
\end{figure}

When evaluating models on datasets with the increased minimum number of symptoms expressed per condition, we can observe a gradual increase in the performance of the models as can be seen in Figure~\ref{figure: inc-symptoms} and Figure~\ref{figure: inc-symptoms-nlice} for the SymCat and NLICE datasets, respectively. As for the SymCat-based dataset, with 5 symptoms per condition, the Naive Bayes model achieved a Top-1 accuracy score of 83.4\% and a precision of 87.6\%. 
Similar improvements are reported for the Random Forest with a Top-1 accuracy of 80.2\% and a precision of 84.2\%. 
Increasing the number of symptoms also contributes to Top-1 accuracy and precision improvements in the NLICE-based dataset. Remarkably, for the Random Forest model, and with a minimum of 5 symptoms per condition, the Top-1 accuracy and precision could achieve almost 100\% and 98\% in the NLICE-based dataset, respectively. This highlights the importance of the NLICE symptom discovery in the differential diagnosis process.

\subsection{Perturbing the condition-symptom probabilities}
Table \ref{table: perturbed-res} shows the Top-1 accuracy performance of the Naive Bayes and Random Forest models on the perturbed data for SymCat and NLICE data. As expected, the table shows that there is a general trend of degrading accuracy with increasing perturbed probabilities. At 70\% perturbation, this degradation can become significant for SymCat data and Naive Bayes on NLICE data. Interestingly, however, the accuracy of the Random Forest model trained on NLICE data is only degraded by 0.5 percentage points for a large 70\% perturbation. This indicates the stability of the predictive capabilities of sophisticated models such as Random Forest combined with the expressiveness of NLICE data.  

\begin{table}[htbp]
\caption{Top-1 accuracy performance on the SymCat and NLICE perturbed dataset (NB = Naive Bayes, RF = Random Forest)}
\centering
\begin{tabular}{|l|ll|ll|}
\hline
\multirow{2}{*}{Dataset} & \multicolumn{2}{c|}{SymCat}        & \multicolumn{2}{c|}{NLICE}         \\ \cline{2-5} 
                         & \multicolumn{1}{l|}{NB}    & RF    & \multicolumn{1}{l|}{NB}    & RF    \\ \hline
Baseline                 & \multicolumn{1}{l|}{0.588} & 0.571 & \multicolumn{1}{l|}{0.820} & 0.820 \\ \hline
Perturbed-10\%           & \multicolumn{1}{l|}{0.594} & 0.580 & \multicolumn{1}{l|}{0.743} & 0.895 \\ \hline
Perturbed-20\%           & \multicolumn{1}{l|}{0.600} & 0.589 & \multicolumn{1}{l|}{0.753} & 0.880 \\ \hline
Perturbed-30\%           & \multicolumn{1}{l|}{0.549} & 0.533 & \multicolumn{1}{l|}{0.730} & 0.864 \\ \hline
Perturbed-50\%           & \multicolumn{1}{l|}{0.696} & 0.679 & \multicolumn{1}{l|}{0.682} & 0.834 \\ \hline
Perturbed-70\%           & \multicolumn{1}{l|}{0.441} & 0.407 & \multicolumn{1}{l|}{0.658} & 0.815 \\ \hline
\end{tabular}
\label{table: perturbed-res}
\end{table}

\subsection{Injected symptoms}
Table~\ref{table: injected-symptoms} shows the model performance on datasets with injected symptoms. In this case, there is a noticeable reduction in performance of the SymCat models even when the symptoms are injected with minimal expression probability. Furthermore, the SymCat models performance is significantly degraded when injected with maximum expression probability, which intuitively can be thought of as the equivalent of making the injected symptoms more relevant to the diagnosis of the condition and thus representing a vary large deviation from the distribution learned by the models. NLICE performance in injected datasets is comparatively more stable than SymCat performance. Even when symptoms with the maximum expression probability are injected, the Top-1 accuracy of the Naive Bayes model trained on the NLICE dataset is degraded by only 5.1 percentage points, indicating that NLICE models could still learn more informative attributes of conditions represented by real-world datasets.

\begin{table*}[tb]
\caption{Model performance on symptom injected SymCat and NLICE datasets (NB = Naive Bayes, RF = Random Forest)}
\centering
\begin{tabular}{|l|llll|llll|llll|}
\hline
\multirow{3}{*}{Dataset} &
  \multicolumn{4}{c|}{Top-1} &
  \multicolumn{4}{c|}{Precision} &
  \multicolumn{4}{c|}{Top-5} \\ \cline{2-13} 
 &
  \multicolumn{2}{l|}{SymCat} &
  \multicolumn{2}{l|}{NLICE} &
  \multicolumn{2}{l|}{SymCat} &
  \multicolumn{2}{l|}{NLICE} &
  \multicolumn{2}{l|}{SymCat} &
  \multicolumn{2}{l|}{NLICE} \\ \cline{2-13} 
 &
  \multicolumn{1}{l|}{NB} &
  \multicolumn{1}{l|}{RF} &
  \multicolumn{1}{l|}{NB} &
  RF &
  \multicolumn{1}{l|}{NB} &
  \multicolumn{1}{l|}{RF} &
  \multicolumn{1}{l|}{NB} &
  RF &
  \multicolumn{1}{l|}{NB} &
  \multicolumn{1}{l|}{RF} &
  \multicolumn{1}{l|}{NB} &
  RF \\ \hline
Baseline &
  \multicolumn{1}{l|}{0.588} &
  \multicolumn{1}{l|}{0.571} &
  \multicolumn{1}{l|}{0.820} &
  0.974 &
  \multicolumn{1}{l|}{0.633} &
  \multicolumn{1}{l|}{0.612} &
  \multicolumn{1}{l|}{0.845} &
  0.975 &
  \multicolumn{1}{l|}{0.853} &
  \multicolumn{1}{l|}{0.845} &
  \multicolumn{1}{l|}{0.975} &
  0.999 \\ \hline
Min Injected &
  \multicolumn{1}{l|}{0.480} &
  \multicolumn{1}{l|}{0.451} &
  \multicolumn{1}{l|}{0.767} &
  0.907 &
  \multicolumn{1}{l|}{0.515} &
  \multicolumn{1}{l|}{0.478} &
  \multicolumn{1}{l|}{0.822} &
  0.913 &
  \multicolumn{1}{l|}{0.754} &
  \multicolumn{1}{l|}{0.743} &
  \multicolumn{1}{l|}{1.000} &
  0.991 \\ \hline
Mean Injected &
  \multicolumn{1}{l|}{0.312} &
  \multicolumn{1}{l|}{0.286} &
  \multicolumn{1}{l|}{0.792} &
  0.895 &
  \multicolumn{1}{l|}{0.380} &
  \multicolumn{1}{l|}{0.340} &
  \multicolumn{1}{l|}{0.832} &
  0.895 &
  \multicolumn{1}{l|}{0.563} &
  \multicolumn{1}{l|}{0.560} &
  \multicolumn{1}{l|}{0.992} &
  0.986 \\ \hline
Max Injected &
  \multicolumn{1}{l|}{0.099} &
  \multicolumn{1}{l|}{0.100} &
  \multicolumn{1}{l|}{0.769} &
  0.730 &
  \multicolumn{1}{l|}{0.207} &
  \multicolumn{1}{l|}{0.177} &
  \multicolumn{1}{l|}{0.783} &
  0.829 &
  \multicolumn{1}{l|}{0.234} &
  \multicolumn{1}{l|}{0.271} &
  \multicolumn{1}{l|}{0.966} &
  0.978 \\ \hline
\end{tabular}
\label{table: injected-symptoms}
\end{table*}

\section{Conclusion}

In this paper, we proposed a systematic approach to constructing synthetic patient records. As has been stated earlier, symptom information is not always enough to make a proper diagnosis, hence additional information regarding the nature, location, chronology, etc, of the symptom, can be gathered from medical literature to increase the available information to the models. We collected additional symptom information for selected conditions. For standardization of the symptom expression, we proposed a novel symptom modeling approach called NLICE and integrated this symptom modeling approach with the Synthea simulator. The analysis demonstrated the suitability of these datasets for using ML models (e.g., Naive Bayes and Random Forest) to the task of estimating a differential diagnosis given a patient's symptoms and demography. We trained Naive Bayes and Random Forest models in both datasets and demonstrated their suitability. The feasibility of using a confidence threshold to filter out the most likely incorrect predictions in a deployed setting was also demonstrated. Qualitatively, models trained in the SymCat-based dataset struggle to distinguish conditions that share similar symptoms. In contrast, this problem did not occur in the models trained in the NLICE-based dataset, which also highlights the importance of introducing additional features and information on symptoms.

Future research directions include the following. We will be working to expanding the number of conditions described by our NLICE data generator. In addition, given the complexity of symptom-disease relationships, we need to explore more powerful ML modeling techniques like autoencoders~\cite{autoencoder-emr}. 
The NLICE code is open sourced at \url{https://github.com/guozhuoran918/NLICE}.

\section*{Acknowledgment}

This research was performed with the support of the EFRO Werk!Werkt project no.\ KVW-00383 and the Eureka Xecs TASTI project no.\ 2022005.  

\bibliographystyle{IEEEtranN} 
\footnotesize
\bibliography{reference.bib}

\end{document}